\documentclass{article}

\usepackage[final]{neurips_mlsb_2020}

\usepackage[utf8]{inputenc} 
\usepackage[T1]{fontenc}    
\usepackage{hyperref}       
\usepackage{url}            
\usepackage{booktabs}       
\usepackage{amsfonts}       
\usepackage{nicefrac}       
\usepackage{microtype}      

\usepackage{times}  
\usepackage{helvet} 
\usepackage{courier}  
\usepackage{graphicx} 
\usepackage{caption} 
\usepackage{enumitem}
\usepackage{multirow}
\usepackage{array}

\usepackage{soul}
\usepackage{subfigure}
\usepackage{amsmath,amsfonts,amssymb,amsthm}
\usepackage{algorithm}
\usepackage{algorithmic}
\usepackage{bibunits}

\newtheorem{theorem}{Theorem}
\newtheorem{definition}{Definition}

\newtheorem{problem}{Problem}

\DeclareMathOperator*{\concat}{\scalebox{1}[1.0]{$\parallel$}}
\newcommand{\angstrom}{\text{\normalfont\AA}}

\title{Molecular Mechanics-Driven Graph Neural Network with Multiplex Graph for Molecular Structures}

\author{%
  Shuo Zhang$^{1,2}$, Yang Liu$^{2}$, Lei Xie$^{1,2,3}$ \\
  \texttt{\{sz780, yl1708, lei.xie\}@hunter.cuny.edu} \\
  $^1$Ph.D. Program in Computer Science, The Graduate Center, The City University of New York \\
  $^2$Department of Computer Science, Hunter College, The City University of New York \\
  $^3$Helen \& Robert Appel Alzheimer’s Disease Research Institute, \\ Feil Family Brain \& Mind Research Institute, \\ Weill Cornell Medicine, Cornell University \\
}

\begin{document}

\maketitle

\begin{abstract}
The prediction of physicochemical properties from molecular structures is a crucial task for artificial intelligence aided molecular design. A growing number of Graph Neural Networks (GNNs) have been proposed to address this challenge. These models improve their expressive power by incorporating auxiliary information in molecules while inevitably increase their computational complexity. In this work, we aim to design a GNN which is both powerful and efficient for molecule structures. To achieve such goal, we propose a molecular mechanics-driven approach by first representing each molecule as a two-layer multiplex graph, where one layer contains only local connections that mainly capture the covalent interactions and another layer contains global connections that can simulate non-covalent interactions. Then for each layer, a corresponding message passing module is proposed to balance the trade-off of expression power and computational complexity. Based on these two modules, we build Multiplex Molecular Graph Neural Network (MXMNet). When validated by the QM9 dataset for small molecules and PDBBind dataset for large protein-ligand complexes, MXMNet achieves superior results to the existing state-of-the-art models under restricted resources.
\end{abstract}

\begin{bibunit}[unsrt]
\section{Introduction}\label{Intro}
Human society benefits greatly from the discovery and design of new molecules with desired properties, from COVID-19 vaccines to solar cells. Artificial intelligence (AI) plays an increasingly important role in accelerating the molecular discovery process. One of the crucial tasks in AI-assisted molecular design is to predict the physicochemical properties of molecules from their structures. In recent years, many machine learning techniques have been proposed for the representational learning of molecules to reduce the computational cost involved in quantum chemistry calculations (DFT) and molecular dynamics simulations (MD)~\cite{chen2018rise}. Among those methods, Graph Neural Networks (GNNs) have shown superior performance by treating the molecule as a graph and performing message passing scheme on it~\cite{wu2020comprehensive}. 

To better model the interactions in molecules and increase the expressive power of methods, previous GNNs have adopted auxiliary information such as chemical properties, pairwise distances between atoms, and angular information~\cite{duvenaud2015convolutional,kearnes2016molecular,gilmer2017neural,schutt2018schnetpack,unke2019physnet,klicpera_dimenet_2020,shui2020heterogeneous}. However, adopting such information in GNNs will inevitably increase the computational complexity. For example, when passing messages on a molecular graph that has $N$ nodes with an average of $k$ nearest neighbors for each node, $O(Nk^2)$ or $O(N^3)$ messages are required in the worst case for the previous state-of-the-art GNNs~\cite{klicpera_dimenet_2020,shui2020heterogeneous} to capture the angular information. With restricted memory resources, those GNNs could exhibit limited expressive power or even fail when applied to macromolecules like proteins or RNAs. 

To address the limitation, we propose a novel GNN that is both powerful and efficient. Inspired by molecular mechanics methods~\cite{schlick2010molecular}, we use the angular information to model only the local connections to avoid using expensive computations on all connections. Besides, we divide the molecular interactions into two categories: local and global. Then a two-layer multiplex graph $G=\{G_l, G_g\}$ is constructed for a molecule. In $G$, the local layer $G_l$ only contains the local connections that mainly capture covalent interactions, and the global layer $G_g$ contains the global connections that cover non-covalent interactions. With the multiplex molecular graphs, we then design Multiplex Molecular (MXM) module that contains a novel angle-aware message passing operated on $G_l$ and an efficient message passing operated on $G_g$. Note that the MXM module reduces the computational complexity by avoiding capturing the angular information in nonlocal interactions. Finally, we construct the \textit{Multiplex Molecular Graph Neural Network} (MXMNet) for the representation learning of molecules.

\begin{figure}[t]
\centering
\vspace{-0.5em}
  \centering
  \includegraphics[scale=0.15]{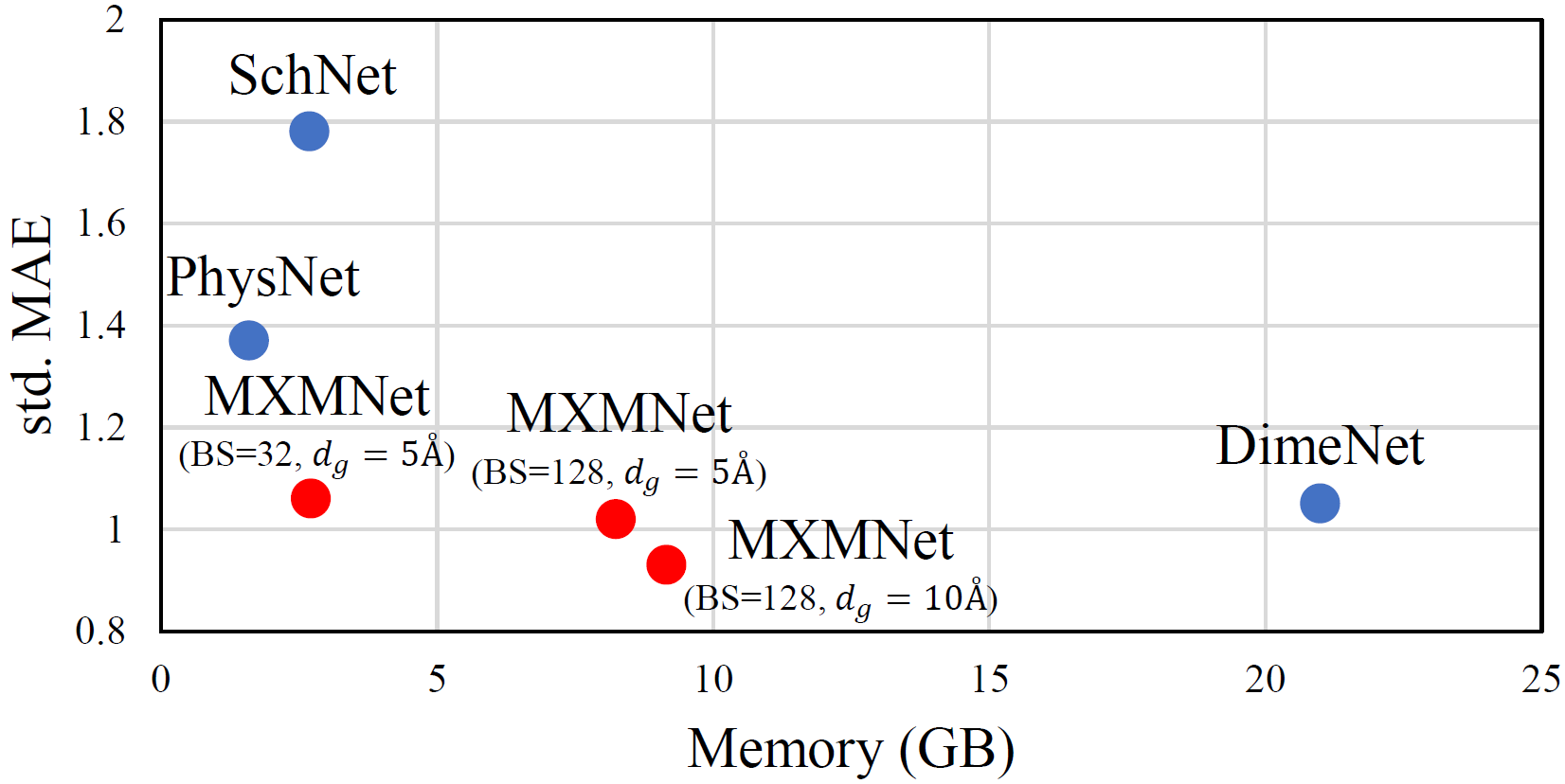}
  \caption{std. MAE vs. memory consumption on QM9 dataset~\cite{ramakrishnan2014quantum}. When compared with Schnet~\cite{schutt2018schnetpack}, PhysNet~\cite{unke2019physnet} and DimeNet~\cite{klicpera_dimenet_2020}, MXMNet gets the state-of-the-art performance and is memory-efficient.}
  \label{fig:efficiency}
\vspace{-0.5em}
\end{figure}

To empirically evaluate the power and efficiency of MXMNet, we conduct experiments on a small molecules dataset QM9~\cite{ramakrishnan2014quantum} and a protein-ligand complexes dataset PDBBind~\cite{wang2004pdbbind}. On both datasets, our model can outperform the baseline models. Regarding the efficiency, our model requires significantly less memory than the previous state-of-the-art model~\cite{klicpera_dimenet_2020} as shown in Figure~\ref{fig:efficiency} and achieves a training speedup of 260$\%$. The main contributions of our work are as follows:
\begin{itemize}
\item We propose a molecular mechanics-driven approach to represent the molecule by using a two-layer multiplex graph, where one layer contains local connections and another layer contains global connections.
\item We propose Multiplex Molecular (MXM) module which performs the message passing on the whole multiplex graph. The MXM module captures the global pairwise distances and local angles to be both powerful and efficient.
\item We propose Multiplex Molecular Graph Neural Network (MXMNet) based on the MXM module. Experiments on benchmark datasets validate that MXMNet achieves state-of-the-art performance and is efficient.
\end{itemize}

\section{Related Work}
\paragraph{GNNs for Molecules.}
To learn the representations of graph-structured data using neural networks, Graph Neural Networks (GNNs) have been proposed~\cite{duvenaud2015convolutional,niepert2016learning,kipf2017semi} and attracted growing interests. Due to the superior performance achieved by GNNs in various tasks, researchers began to apply GNNs for predicting various properties of molecules. Initial works treat the chemical bonds in molecules as edges and atoms as nodes to create graphs for molecules~\cite{duvenaud2015convolutional,kearnes2016molecular,gilmer2017neural}. These GNNs also integrate many hand-picked chemical features to improve performance. However, they do not take account of the 3-dimensional structure of molecules, which is critical for many physiochemical properties of molecules. Thus later works~\cite{schutt2017quantum,schutt2018schnetpack,chen2019graph,unke2019physnet} turn to take the atomic positions into consideration and use interatomic distances to create the edges as well as edge features between atoms. Usually, a cutoff distance is used to find the neighbors in molecules instead of creating a complete graph to reduce the computational complexity and overfitting. However, the setting of cutoff sometimes can lead the GNNs to fail to distinguish certain molecules~\cite{klicpera_dimenet_2020}. To solve this issue, angular information is further used in GNNs to achieve higher expressive power~\cite{klicpera_dimenet_2020, shui2020heterogeneous}. However, those angle-aware GNNs have significantly higher time and space complexity than the previous works. They are not scalable to macromolecules or large-batch learning.

\paragraph{Multiplex Graph.}
The multiplex graph (a.k.a multi-view graph) consists of multiple types of edges among a set of nodes. Informally, it can be considered as a collection of graphs, where each type of edges with the same set of nodes forms a graph or a layer. To get the representation of each node, both intra-layer relationships and cross-layer relationships have to be addressed properly. In practice, various methods have been proposed to learn the embedding of the multiplex graph~\cite{liu2017principled,zhang2018scalable,schlichtkrull2018modeling,cen2019representation,ma2019multi} and the multiplex graph can be applied in many fields~\cite{kao2018layer,lee2020heterogeneous,wang2020abstract}. For the representation learning on molecules, previous work~\cite{shi2020graph} implicitly represents molecular graphs as multiplex graphs and passes messages according to the edge types. In this work, we explicitly represent molecules as multiplex graphs based on the geometric information in molecules. Moreover, we propose different message passing schemes for different layers in the multiplex graph.

\section{Preliminaries}
In this section, we will introduce the preliminaries about our work. We first introduce the main notations used in this paper. Let $G = ( V , E )$ be a graph with $N=|V|$ nodes and $M = |E|$ edges. The nearest neighbors of node $i$ are defined as $\mathcal { N } ( i ) = \{ j | d ( i , j ) = 1 \}$, where $d ( i , j )$ is the shortest distance between node $i$ and $j$. The average number of the nearest neighbors of each node is $k = 2M / N$. In the later formulations, we will use $\boldsymbol{h}_{i}$ as the embedding of node $i$, $\boldsymbol{e}_{j i}$ as the edge embedding between node $i$ and $j$, which embeds the pairwise distance, $\boldsymbol{m}_{ji}$ as the message being sent from node $j$ to node $i$ in the message passing scheme~\cite{gilmer2017neural}, $\mathrm{MLP}$ as the multi-layer perceptron, $\concat$ as the concatenation operation, $\odot$ as the element wise production and $\boldsymbol{W}$ as the weight matrix. Next we provide the definition of a multiplex graph:

\begin{definition} \label{def:Multiplex} \textbf{Multiplex Graph.} 
A multiplex graph can be defined as an $L+1$-tuple $G = (V, E^1, \ldots, E^L)$ where $V$ is the set of nodes and for each $l\in\{1, 2, \ldots, L\},$ $E^l$ is the set of edges in type $l$ that between pairs of nodes in $V$. By defining the graph $G^l = (V,E^l)$ which is also called a plex or a layer, the multiplex graph can be seen as the set of graphs $G = \{G^1, G^2, ..., G^L\}$.
\end{definition}

Now we introduce the message passing scheme~\cite{gilmer2017neural} which is a general graph convolution used in spatial-based GNNs~\cite{wu2020comprehensive}:

\begin{definition} \label{def:MP} \textbf{Message Passing.} 
Given a graph $G$, the node feature of each node $i$ is $\boldsymbol{x}_i$, and the edge feature for each node pair $j$ and $i$ is $\boldsymbol{e}_{ji}$. The message passing scheme iteratively updates the node embedding $\boldsymbol{h}$ using the following functions:
\begin{align}
\boldsymbol{m}_{ji}^{t} = f_{\text {m}}(\boldsymbol{h}_{i}^{t-1}, \boldsymbol{h}_{j}^{t-1}, \boldsymbol{e}_{ji}), \ \ \ \ \ \boldsymbol{h}_{i}^{t} = f_{\text {u}}(\boldsymbol{h}_{i}^{t-1}, \sum\nolimits_{j \in \mathcal{N}(i)} \boldsymbol{m}_{ji}^{t}),
\nonumber
\end{align}
where the superscript $t$ denotes the $t$-step iteration, $\boldsymbol{h}_{i}^{0} =\boldsymbol{x}_i$, the $f_{\text {m}}$ and $f_{\text {u}}$ are learnable functions.
\end{definition}

In recent works~\cite{klicpera_dimenet_2020,shui2020heterogeneous}, the message passing scheme has been modified to capture the angular information in a 3D molecular graph $G = (V, E)$ with $N$ nodes and their Cartesian coordinates $\boldsymbol{r}=\left\{\boldsymbol{r}_{1}, \ldots, \boldsymbol{r}_{N}\right\}$, where $\boldsymbol{r}_{i} \in \mathbb{R}^3$ is the position of node $i$. To analyze their computational complexity, we start from the number of angles in $G$ to be captured:

\begin{theorem} \label{theo}
Given a 3D molecular graph $G$, each pair of adjacent edges that share a common node can define an angle in $G$. There are $O(Nk^2)$ angles in $G$, where $N$ is the number of nodes and $k$ is the average number of nearest neighbors for each node.
\end{theorem}

The proof is straightforward: For each node in $G$, there is an average of $k$ edges connected to it. Those $k$ edges can define $(k(k-1))/2$ angles. Thus in total, we have $O(Nk^2)$ angles in $G$. To capture those angles in message passing scheme, there is at least one message being used to contain each angle in recent approaches~\cite{klicpera_dimenet_2020,shui2020heterogeneous}. Thus the computational complexity of those models is at least $O(Nk^2)$ for each graph in an operation.

Finally we present the problem investigated in this work:

\begin{problem}  \label{problem} \textbf{Molecular Properties Prediction.}
Given a molecule with $N$ atoms and their atomic numbers $\boldsymbol{Z}=\left\{Z_{1}, \ldots, Z_{N}\right\}$ and Cartesian coordinates $\boldsymbol{r}=\left\{\boldsymbol{r}_{1}, \ldots, \boldsymbol{r}_{N}\right\}$, the problem of molecular properties prediction is to predict the target property $t \in \mathbb{R}$ of the molecule. The regression goal is to find a function $f: \{\boldsymbol{Z}, \boldsymbol{r}\} \rightarrow \mathbb{R}$. Sometimes with auxiliary chemical information $\boldsymbol{\Theta}$, the goal function is $f: \{\boldsymbol{Z}, \boldsymbol{r}, \boldsymbol{\Theta}\} \rightarrow \mathbb{R}$.
\end{problem}

\section{Approach}
In this section, we introduce our molecular mechanics-driven approach including the multiplex molecular graphs, the Multiplex Molecular (MXM) module, and the Multiplex Molecular Graph Neural Network (MXMNet).

\begin{figure*}[t]
    \vspace{-0.5em}
    \centering
	\includegraphics[width=\textwidth]{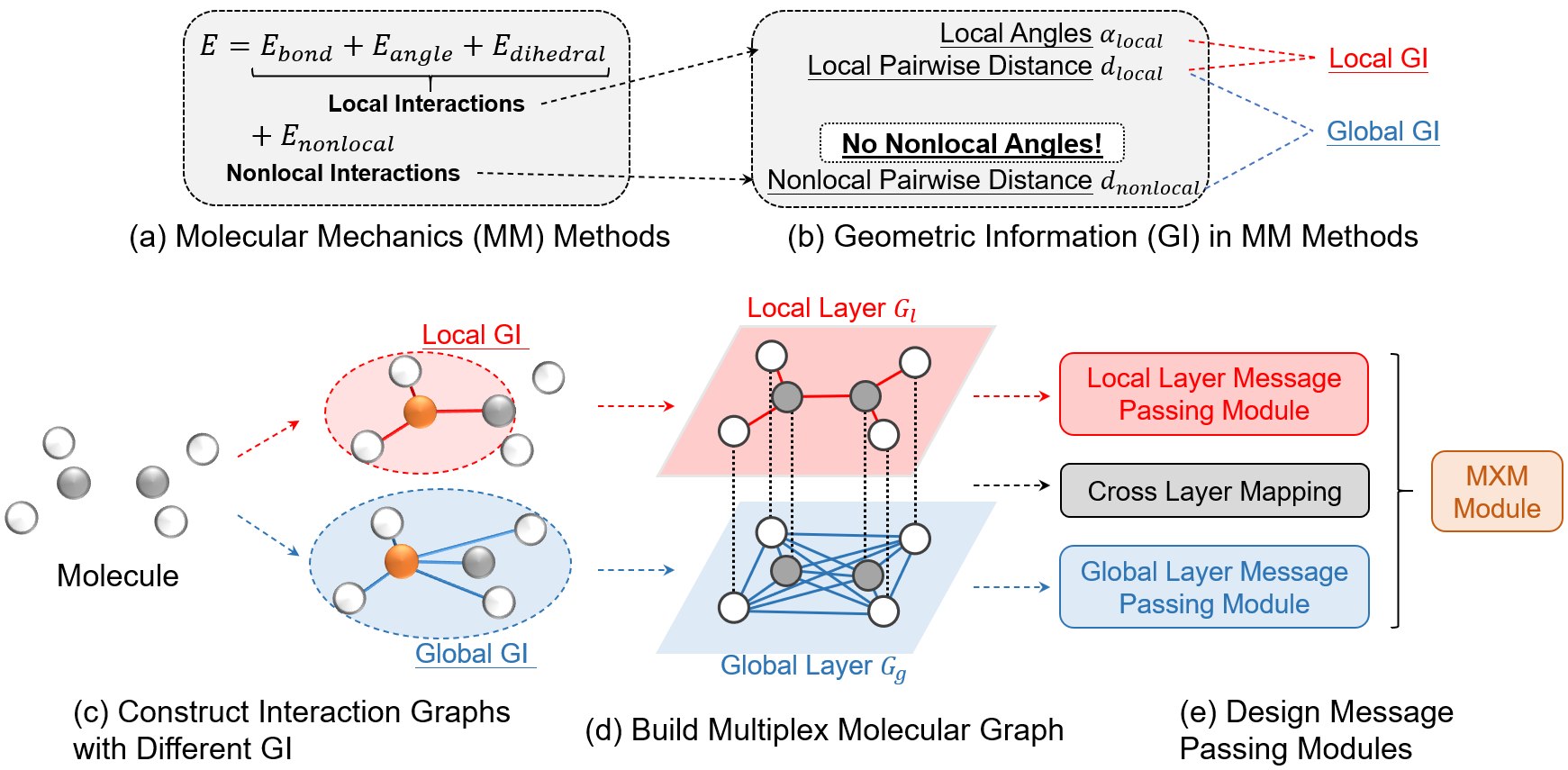}
	\caption{\label{fig:Pipeline}Illustration of our molecular mechanics-driven approach. (a) The molecular mechanics (MM) methods. (b) The geometric information (GI) used in MM methods does not contain the angles in nonlocal interactions. We further group the GI into Local GI and Global GI. (c) Given a 3D molecule, interaction graphs are constructed with different GI. We show an example of creating edges around an orange node. (d) The resulting interaction graphs are used to build a multiplex molecular graph $G$. (e) With $G$, we design message passing modules to update the node embeddings hierarchically and efficiently.}
	\vspace{-0.5em}
\end{figure*}

\subsection{Multiplex Molecular Graphs} \label{sec:multiplex}
In molecular mechanics methods~\cite{schlick2010molecular}, the molecular energy $E$ is modeled as $E=E_{\text {local}}+E_{\text {nonlocal}}$ (see Figure~\ref{fig:Pipeline}(a)), where $E_{\text {local}}=E_{\text {bond}}+E_{\text {angle}}+E_{\text {dihedral}}$ models the local, covalent interactions including $E_{\text {bond}}$ that depends on bond lengths, $E_{\text {angle}}$ on bond angles, and $E_{\text {dihedral}}$ on the dihedral angles. $E_{\text {nonlocal}}$ models the non-local, non-covalent interactions between atom pairs. When focusing on the geometric information contained in the molecular mechanics method, we will find that the local interactions capture the angles $\alpha _{local}$ and the pairwise distances $d_{local}$ while the nonlocal interactions only capture the pairwise distances $d_{nonlocal}$ (see Figure~\ref{fig:Pipeline}(b)). These inspire us to use the angular information to model \textbf {only} the local interactions instead of all interactions in our model to reduce the computational complexity.

To achieve our goal, we first divide the geometric information (GI) in molecular mechanics methods into two groups: Local GI that contains $\alpha _{local}$ and $d_{local}$, and Global GI that contains $d_{local}$ and $d_{nonlocal}$ (see Figure~\ref{fig:Pipeline}(b)). Given a 3D molecule, we then construct the corresponding interaction graphs that contain different GI (see Figure~\ref{fig:Pipeline}(c)). For the local GI, we can create edges by using either chemical bonds or finding the neighbors of each node within a small cutoff distance depending on the task being investigated. For the global GI, we create the edges by defining the neighbors of each node within a relatively large cutoff distance. With the interaction graphs, we treat them as layers to build a multiplex molecular graph $G = \{G_{l}, G_{g}\}$, which consists of a \textbf {local layer} $G_{l}$ and a \textbf {global layer} $G_{g}$ (see Figure~\ref{fig:Pipeline}(d)). The resulting $G$ will be used as the input of our model.

\subsection{Multiplex Molecular (MXM) Module} \label{sec:MXM module}
With the multiplex molecular graph $G$, we propose Multiplex Molecular (MXM) module that uses different rules to update the node embeddings based on the different edges in $G$ (see Figure~\ref{fig:Pipeline}(e)). For $G_{g}$, we propose the \textbf {global layer message passing}. For $G_{l}$, we propose the \textbf {local layer message passing}. To transfer the information between different layers, we use a \textbf {cross layer mapping}. These operations will be introduced as follows in detail.

\paragraph{Global Layer Message Passing Module.}
In this module, the message passing is performed on the global layer, which contains both local and non-local connections. We propose a message passing module that can capture the pairwise distances based on the message passing defined in Definition~\ref{def:MP}. Note that the message passing in Definition~\ref{def:MP} can \textit{only} take the one-hop neighbors of the central node in the aggregation per iteration. Inspired by previous works that demonstrate the power of addressing high-order neighbors in GNNs~\cite{liu2019geniepath,yang2019spagan,abu2019mixhop}, we here propose a message passing that captures up to the two-hop neighbors per iteration. A straightforward way to achieve the goal would be directly aggregating all two-hop neighbors. However, this would require $O(Nk^2)$ messages on the graph per iteration. Instead, we perform the one-hop based message passing twice in each iteration to address the two-hop neighbors. The resulting operation will only need $O(2Nk)$ messages in this way. 

\begin{figure*}[t]
    \vspace{-0.5em}
    \centering
	\includegraphics[width=\textwidth]{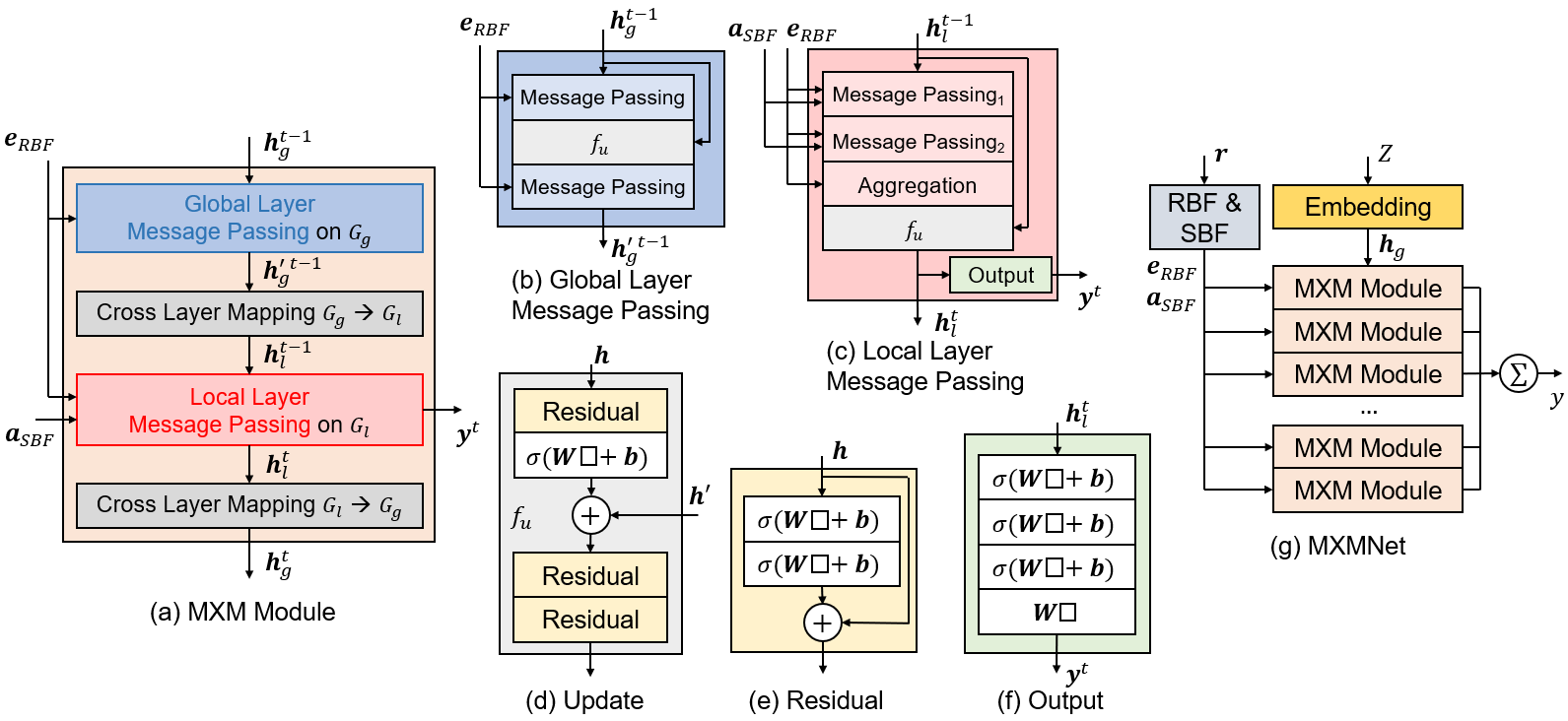}
	\caption{\label{fig:MXMNet}Overview of the architecture of the MXM module and the MXMNet. In the illustrations, $\sigma$ denotes the non-linear transformation, $\square$ denotes the input for the layer.}
	\vspace{-0.5em}
\end{figure*}

As illustrated in Figure~\ref{fig:MXMNet}(b), our \textbf {global layer message passing} module consists of two \textit{identical} message passing operations that can capture the pairwise distance information $\boldsymbol{e}$. Each message passing operation is formulated as follows:
\begin{align}
\boldsymbol{m}_{ji} &= \mathrm{MLP}([\boldsymbol{h}_{j}^{input} \concat \boldsymbol{h}_{i}^{input} \concat \boldsymbol{e}_{j i}]) \odot (\boldsymbol{e}_{j i}\boldsymbol{W}), \\
\boldsymbol{h}_{i}^{output} &= \boldsymbol{h}_{i}^{input} + \sum\nolimits_{j \in \mathcal{N}(i)} \boldsymbol{m}_{ji},
\end{align}
where $i, j \in G_{global}$, the superscripts denote the state of $\boldsymbol{h}$ in the operation. In our global layer message passing, an update function $f_{\text {u}}$ is used between the two message passing operations. We define $f_{\text {u}}$ using multiple residual modules (see Figure~\ref{fig:MXMNet}(d)). Each residual module consists of a two-layer MLP and a skip connection (see Figure~\ref{fig:MXMNet}(e)).

\begin{figure}[t]
    \vspace{-0.5em}
    \centering
	\includegraphics[scale=0.6]{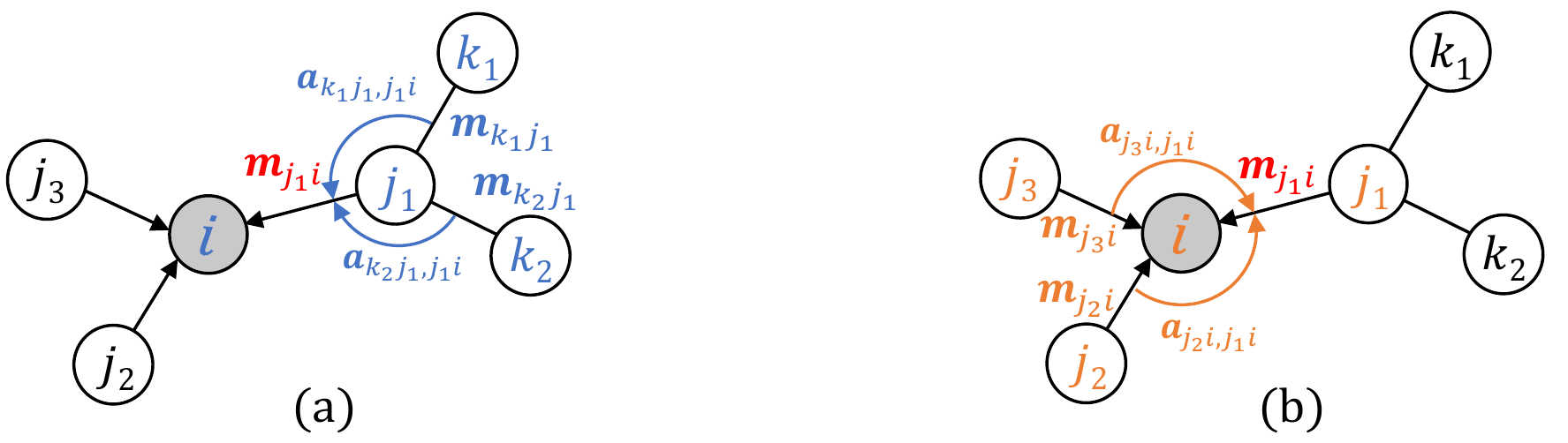}
	\caption{\label{fig:local_message_passing} Illustration of Message Passing 1 and 2 used in the Local Layer Message Passing module. (a) Message Passing 1 can capture the two-hop angles like $\angle i j_1 k_1$ and $\angle i j_1 k_2$ when updating $\boldsymbol{m}_{j_1 i}$. (b) Message Passing 2 can capture the one-hop angles like $\angle j_1 i j_3$ and $\angle j_1 i j_2$ when updating $\boldsymbol{m}_{j_1 i}$.}
	\vspace{-0.5em}
\end{figure}

\paragraph{Local Layer Message Passing Module.} 
In this module that performs message passing on the local layer, we will incorporate both the pairwise distance and angles associated with local interactions. In practice, we propose a message passing that captures up to the two-hop neighbors per iteration. In this way, the edges can define two kinds of angles: The \textbf {two-hop angles} that between the one-hop edges and the two-hop edges ($\angle i j_1 k_1$, $\angle i j_1 k_2$ in Figure~\ref{fig:local_message_passing}). The \textbf {one-hop angles} that only between the one-hop edges ($\angle j_1 i j_2$ and $\angle j_1 i j_3$ in Figure~\ref{fig:local_message_passing}). Our message passing can capture all of those angles. While the previous work~\cite{klicpera_dimenet_2020} only captures the two-hop angles.

In detail, we propose a 3-step message passing scheme to be the \textbf {local layer message passing}: Step 1 contains Message Passing 1 that captures the two-hop angles and related pairwise distances to update edge-level embeddings $\{\boldsymbol{m}_{ji}\}$ (see Figure~\ref{fig:local_message_passing}(a)). Step 2 contains Message Passing 2 that captures the one-hop angles and related pairwise distances to further update $\{\boldsymbol{m}_{ji}\}$ (see Figure~\ref{fig:local_message_passing}(b)). Step 3 finally aggregates $\{\boldsymbol{m}_{ji}\}$ to update the node-level embedding $\boldsymbol{h}_{i}$. These steps in the $t$-th iteration can be formulated as follows:
\begin{align}
\intertext{Step 1: Message Passing 1}
\boldsymbol{m}_{kj}^{t-1} &= \mathrm{MLP_{kj}}([\boldsymbol{h}_{k}^{t-1} \concat \boldsymbol{h}_{j}^{t-1} \concat \boldsymbol{e}_{kj}]) \odot (\boldsymbol{e}_{kj}\boldsymbol{W}_{e1}) \odot \mathrm{MLP_{a1}}(\boldsymbol{a}_{k j, j i}) ,\\
\boldsymbol{m}_{ji}^{t-1} &= \mathrm{MLP_{ji}}([\boldsymbol{h}_{j}^{t-1} \concat \boldsymbol{h}_{i}^{t-1} \concat \boldsymbol{e}_{ji}]) + \sum\nolimits_{k \in \mathcal{N}(j)\setminus\{i\}} \boldsymbol{m}_{kj}^{t-1}, \\
\intertext{Step 2: Message Passing 2} 
\boldsymbol{m}_{j'i}^{'t-1} &= \mathrm{MLP_{j'i}}(\boldsymbol{m}_{j'i}^{t-1}) \odot (\boldsymbol{e}_{j'i}\boldsymbol{W}_{e2}) \odot \mathrm{MLP_{a2}}(\boldsymbol{a}_{j'i, j i}),\\
\boldsymbol{m}_{ji}^{'t-1} &= \mathrm{MLP_{ji}^{'}}(\boldsymbol{m}_{ji}^{t-1}) + \sum\nolimits_{j' \in \mathcal{N}(i)\setminus\{j\}} \boldsymbol{m}_{j'i}^{'t-1}, \\
\intertext{Step 3: Aggregation and Update} 
\boldsymbol{h}_{i}^{t} &= f_{\text {u}}(\sum\nolimits_{j \in \mathcal{N}(i)} \boldsymbol{m}_{ji}^{'t-1}\odot (\boldsymbol{e}_{ji}\boldsymbol{W}_{e3})) ,
\end{align}
where $i, j, k \in G_{local}$, $\boldsymbol{a}_{k j, j i}$ is the feature for angle $\alpha_{k j, j i}=\angle \boldsymbol{h}_{k} \boldsymbol{h}_{j} \boldsymbol{h}_{i}$. We define $f_{\text {u}}$ using the same form as in the global layer message passing. These steps need $O(2Nk^2 + Nk)$ messages in total.

Figure~\ref{fig:MXMNet}(c) illustrates the architecture of the global layer message passing. Note that we also include an Output module (see Figure~\ref{fig:MXMNet}(c) and (f)), which is used for producing the output when creating the whole GNN model later.

\paragraph{Cross Layer Mapping.}
After having the message passing modules for the local and global layer, we further use a cross layer mapping function $f_{cross}$ to address the connections between the same nodes across different layers in a multiplex molecular graph (see Figure~\ref{fig:Pipeline}(d)). 

The cross layer mapping function $f_{cross}$ takes either the node embeddings $\{\boldsymbol{h}_g\}$ in the global layer or the node embeddings $\{\boldsymbol{h}_l\}$ in the local layer as input, and maps them to replace the node embeddings in the other layer (see Figure~\ref{fig:MXMNet}(a)): 
\begin{align}
\boldsymbol{h}_l = f_{cross}(\boldsymbol{h}_g)  \ \ \ \text{or}\ \ \  \boldsymbol{h}_g = f'_{cross}(\boldsymbol{h}_l), 
\end{align}
where $g\in G_{global}, l\in G_{local}$, the $f_{cross}$ and $f'_{cross}$ are learnable functions. In practice, we use multi-layer perceptrons to be $f_{cross}$ and $f'_{cross}$. Each of them needs $O(N)$ messages being updated.

\subsection{Multiplex Molecular Graph Neural Network (MXMNet)} \label{sec:MXMNet}
With MXM module, we build Multiplex Molecular Graph Neural Network (MXMNet) for the prediction of molecular properties as shown in Figure~\ref{fig:MXMNet}(g). In the \textbf {Embedding module}, the atomic numbers $Z$ are represented with randomly initialized, trainable embeddings to be the input node embeddings. In the \textbf {RBF \& SBF module}, the Cartesian coordinates $\boldsymbol{r}$ of atoms are used to compute the pairwise distances and angles. We use the basis functions proposed in~\cite{klicpera_dimenet_2020} to construct the representations of $\boldsymbol{e}_{RBF}$ and $\boldsymbol{a}_{SBF}$. Then we stack \textbf {MXM modules} to perform message passings. In each MXM module, we use an \textbf {Output module} to get the node-level output. The final prediction $y$ is computed by summing all outputs together among all nodes and all layers.

\subsection{Expressive Power and Complexity of MXMNet}
\paragraph{Expressive Power.}
We analyze the expressive power of MXMNet by focusing on the effect of captured geometric information on representing molecular structures. Since MXMNet takes the pairwise distance information in global connections and the angular information in local connections into consideration, it is more powerful than the GNNs that only captures the pairwise distance information~\cite{schutt2017quantum,schutt2018schnetpack,chen2019graph,unke2019physnet}. When compared with the GNNs that captures both the pairwise distance information and angular information in global connections~\cite{klicpera_dimenet_2020, shui2020heterogeneous}, MXMNet theoretically has lower expressive power due to the uncaptured angular information in nonlocal connections. However, note that expressive power does not
directly speak about the generalization ability of GNNs~\cite{zhang2016understanding,xu2018how}, our experiments will empirically show that MXMNet exhibits good generalization ability with state-of-the-art performance.

\paragraph{Computational Complexity.}
To analyze the computational complexity, we focus on the time and space complexity of message passing in MXMNet. We denote the cutoff distance when creating the edges as $d_g$ and $d_l$ in $G_g$ and $G_l$. The average number of the nearest neighbors per node is $k_g$ in $G_g$ and is $k_l$ in $G_l$. For 3D molecules, we have $k_g \propto {d_g}^3$ and $k_l \propto {d_l}^3$. As $d_g > d_l$, we know that $k_g \gg k_l$. As discussed in previous sections, the message passing operations in our MXM module requires the computation of $O(2Nk_g+2N{k_l}^2+Nk_l+2N)$ messages in total. Therefore, MXMNet is much more efficient than the GNNs capturing angular information in global connections~\cite{klicpera_dimenet_2020, shui2020heterogeneous}, which require $O(N{k_g}^2)$ messages.

\section{Experiments}
In our experiments, we evaluate the generalization power as well as the efficiency of our MXMNet on the QM9 dataset for predicting molecular properties and the PDBBind dataset for predicting the protein-ligand binding affinities. Several state-of-the-art baseline models are also included for comparisons.
\subsection{Experimental Setup}
\paragraph{QM9.}
The QM9 dataset is a widely used benchmark for the prediction of physical properties of molecules in
equilibrium~\cite{ramakrishnan2014quantum}. It consists of around 130k small organic molecules with up to 9 heavy atoms (C, O, N, and F). The properties are computed using density functional theory (DFT) calculations. Following~\cite{klicpera_dimenet_2020}, we randomly use 110000 molecules for training, 10000 for validation and the rest for testing. We evaluate the mean absolute error (MAE) of the target properties. To create the multiplex molecular graphs, we use the chemical bonds as the edges in the local layer, and a cutoff distance to create the edges in the global layer.

\paragraph{PDBBind.}
PDBBind is a database of experimentally measured binding affinities for protein-ligand complexes~\cite{wang2004pdbbind}. It contains detailed 3D structures and associated inhibition constants $K_i$ for the complexes. In our experiment, we use the PDBBind 2015 refined subset which contains roughly 4K structures. In each complex, we exclude the protein residues that are more than 6$\angstrom$ from the ligand. Besides, we remove all hydrogen atoms and use the remaining heavy atoms in the structure. The resulting complexes contain around 200 atoms on average. In the experiment, we split the dataset into training, validation, and testing sets by 8:1:1 and perform 10-fold cross-validation. The mean absolute error (MAE) of the binding free energy and the Pearson correlation coefficient (R) of log$K_i$ are reported. To create the multiplex molecular graphs, we use a cutoff distance of $2\angstrom$ in the local layer and $6\angstrom$ in the global layer when defining the edges.

In our experiments, we use the following state-of-the-art models as baselines: SchNet~\cite{schutt2018schnetpack}, PhysNet~\cite{unke2019physnet}, MEGNet-full~\cite{chen2019graph}, Cormorant~\cite{anderson2019cormorant}, MGCN~\cite{lu2019molecular} and DimeNet~\cite{klicpera_dimenet_2020}. On QM9, we use the results reported in the original works for the baselines. On PDBBind, we conduct the experiments based on the corresponding implementations. All of the experiments are done on an NVIDIA Tesla V100 GPU (32 GB). More details of the parameter settings and training setup are included in the appendix.

\subsection{Results on QM9}
On the QM9 dataset, we test the performance of MXMNet under different configurations by changing the batch size BS and the cutoff distance $d_g$ used in the global layer. As reported in Table~\ref{table:QM9}, MXMNet variants get better results than the baselines on 9 targets. We also compute the mean standardized MAE (std. MAE) as used in~\cite{klicpera_dimenet_2020} to evaluate the overall performance of the models. MXMNet (BS=128, $d_g=10\angstrom$) has the lowest std. MAE among all models and decreases the mean std. MAE by 13$\%$ compared to the previous best model DimeNet. The results clearly demonstrate the excellent generalization power of MXMNet.

\begin{table*}[t]
\vspace{-0.5em}
\caption{Comparison of MAEs of targets on QM9 for different models.}
\label{table:QM9}
\centering
\resizebox{\textwidth}{!}{%
\begin{tabular}{cccccccccc}
\toprule
\multirow{3}{*}{Target} & \multirow{3}{*}{SchNet} & \multirow{3}{*}{PhysNet} & \multirow{3}{*}{MEGNet-f} & \multirow{3}{*}{Cormorant} & \multirow{3}{*}{MGCN} & \multirow{3}{*}{DimeNet} & \textbf{MXMNet} & \textbf{MXMNet} & \textbf{MXMNet}\\
&  &  &  &  &  &  & BS=32 & BS=128 & BS=128\\
&  &  &  &  &  &  & $d_g$=5$\angstrom$ & $d_g$=5$\angstrom$ & $d_g$=10$\angstrom$\\
\midrule
$\mu$ (D)          & $\textbf{0.021}$ & 0.0529 & 0.040 & 0.038 & 0.056 & 0.0286 & 0.0396 & 0.0382 & 0.0255\\
$\alpha (a_0^3)$  & 0.124 & 0.0615 & 0.083 & 0.085 & $\textbf{0.030}$ & 0.0469 & 0.0447 & 0.0482 & 0.0465\\
$\epsilon_{\text{HOMO}}$ (meV)  & 47 & 32.9 & 38 & 34 & 42.1 & 27.8 & 24.7 & 23.0 & $\textbf{22.8}$\\
$\epsilon_{\text{LUMO}}$ (meV)  & 39 & 24.7 & 31 & 38 & 57.4 & 19.7 & 19.7 & 19.5 & $\textbf{18.9}$\\
$\Delta\epsilon$ (meV)            & 74 & 42.5 & 61 & 61 & 64.2 & 34.8 & 32.6 & 31.2 & $\textbf{30.6}$\\
$\left\langle R^{2}\right\rangle (a_0^2)$  & 0.158 & 0.765 & 0.265 & 0.961 & 0.11 & 0.331 & 0.512 & 0.506 & $\textbf{0.088}$\\
ZPVE (meV)   & 1.616 & 1.39 & 1.40 & 2.027 & $\textbf{1.12}$ & 1.29 & 1.15 & 1.16 & 1.19\\
$U_0$ (meV)  & 12 & 8.15 & 9 & 22 & 12.9 & 8.02 & $\textbf{5.90}$ & 6.10 & 6.59\\
$U$ (meV)    & 12 & 8.34 & 10 & 21 & 14.4 & 7.89 & $\textbf{5.94}$ & 6.09 & 6.64\\
$H$ (meV)    & 12 & 8.42 & 10 & 21 & 16.2 & 8.11 & $\textbf{6.09}$ & 6.21 & 6.67\\
$G$ (meV)    & 13 & 9.40 & 10 & 20 & 14.6 & 8.98 & $\textbf{7.17}$ & 7.30 & 7.81\\
$c_v (\frac{\mathrm{cal}}{\mathrm{mol} \mathrm{K}})$  & 0.034 & 0.0280 & 0.030 & 0.026 & 0.038 & 0.0249 & $\textbf{0.0224}$ & 0.0228 & 0.0233\\
\midrule
std. MAE ($\%$ ) & 1.78 & 1.37 & 1.57 & 1.61 & 1.89 & 1.05 & 1.06 & 1.02 & $\textbf{0.93}$\\
\bottomrule
\end{tabular}%
}
\vspace{-0.5em}
\end{table*}

\begin{table*}[t]
\vspace{-0.5em}
\caption{Comparison of mean std. MAEs of ablations that only contain parts of the MXM module. MXMNet cannot achieve the stat-of-the-art performance without any part of the MXM module.} \label{table:ablation}
\centering
\newcommand{\tabincell}[2]{\begin{tabular}{@{}#1@{}}#2\end{tabular}}
\small
\begin{tabular}{lcc}
	\toprule
	\multicolumn{2}{c}{Ablation} & $\frac{\text {std. MAE}}{\text {std. MAE of MXMNet}}$  \\
	\midrule
	\multirow{2}{*}{\tabincell{c}{Only Global Layer \\ Message Passing}}   & One MP operation & 116\%\\
     & Two MP operations & 110\%\\
    \midrule
     \multirow{3}{*}{\tabincell{c}{Only Local Layer \\ Message Passing}}   & Step 1, 3 & 266\%\\
       & Step 2, 3 & 244\%\\
       & Step 1, 2, 3 & 224\%\\
	\bottomrule
\end{tabular}
\vspace{-0.5em}
\end{table*}

\paragraph{Ablation Study.}
By comparing the results between MXMNet (BS=32, $d_g=5\angstrom$) and  MXMNet (BS=128, $d_g=5\angstrom$), we find that the effect of batch size on the performance is small. With a relatively large batch size (128), the overall performance is slightly better than using a small batch size (32). Moreover, we can benefit from the large batch to achieve faster training.

To investigate the effect of $d_g$ on the performance, we compare the results of the MXMNet variants that using different $d_g$ in Table~\ref{table:QM9}. When using $d_g=5\angstrom$, MXMNet can get better results than using $d_g=10\angstrom$ on the targets ZPVE, $U_0$, $U$, $H$, $G$ and $c_v$. This suggests that those properties benefit more from modeling a limited range of interactions rather than simply increasing the interaction range. While for the targets $\mu$, $\epsilon_{\text{HOMO}}$, $\epsilon_{\text{LUMO}}$, $\Delta\epsilon$ and $\left\langle R^{2}\right\rangle$, the performance of MXMNet can be improved by using a larger $d_g=10\angstrom$ that helps to capture longer range interactions. Therefore, in practice, it is recommended to use different $d_g$ for predicting different properties.

To further test whether our proposed two message passing modules (local layer and global layer) will both contribute to the success of MXMNet, we conduct experiments by using only one of the two modules or even parts of a module. Table~\ref{table:ablation} shows that all the ablations will decrease the performance of MXMNet. These validate that both of the two message passing modules contribute to the power of MXMNet. Besides, when only using the global layer message passing module, the ablation with only one message passing performs worse than the ablation with two message passings, which shows the effectiveness of capturing the two-hop neighbors. When only using the local layer message passing module, the mean std. MAE increases significantly compared to the original MXMNet, suggesting that the local connections are not adequate for the task. The results also validate the necessity to capture both one-hop angles and two-hop angles: The ablations with either one kind of them perform worse than the ablation with all of them.

\paragraph{Efficiency Evaluation.}
To evaluate the space and time efficiency of MXMNet, we first compare the memory consumption during the training on QM9 for SchNet, PhysNet, DimeNet, and MXMNet. For the baselines, the model configurations are the same as those in their original papers. As illustrated in Figure~\ref{fig:efficiency}, all of the three MXMNet variants use a much smaller memory than DimeNet. For SchNet and PhysNet that consume less memory than MXMNet, they perform worse than MXMNet with higher mean std. MAEs. Then for time efficiency, we focus on the total training time. Note that the total training time is affected by all operations in the models and different models need different computational time for passing one message. Thus a smaller number of messages being passed in a GNN does not guarantee a shorter training time. Instead, we find that the batch size significantly affects the training time: MXMNet can benefit from large batch training with BS=128 to achieve a speedup of the training to 2.6$\times$ against DimeNet that can only use BS=32 on our GPU.

\begin{table*}[t]
\caption{Results of the Pearson correlation coefficient $R$ and MAEs of different models on PDBBind. '-' denotes that the model raises out-of-memory issue.}
\label{table:PDBBind}
\small
\centering
\begin{tabular}{lcc}
	\toprule
	Model & Pearson $R$ & MAE  \\
	\midrule
	SchNet & 0.601$\pm$0.037 & 1.892$\pm$0.071\\
	PhysNet & 0.614$\pm$0.034 & 1.881$\pm$0.065\\
	DimeNet & - & -\\
	\textbf{MXMNet} & {\bf 0.664 $\pm$ 0.024} & {\bf 1.733 $\pm$ 0.089}\\
	\bottomrule
\end{tabular}	
\vspace{-0.5em}
\end{table*}

\subsection{Results on PDBBind}
On the PDBBind dataset with much larger molecules than those in QM9, the training of MXMNet is still able to be performed on our GPU. With the same model configuration, DimeNet will raise the out-of-memory error. As shown in Table~\ref{table:PDBBind}, when compared with SchNet and PhysNet that do not have the memory issue, MXMNet outperforms them significantly with a higher Pearson $R$ and a lower MAE. Those results validate that our model is both powerful and memory-efficient to be used for macromolecules.

\section{Conclusion}
In this paper, we propose a powerful and efficient GNN, MXMNet, for predicting the properties of molecules. Our model can significantly improve both expressive power and memory efficiency of GNNs for molecules. The novelty of MXMNet lies in its representation of molecules as a multiplex graph that is rooted in molecular mechanics. Experiments on QM9 and PDBBind datasets have successfully demonstrated the power and efficiency of MXMNet compared with the state-of-the-art baselines. In future work, it would be interesting to address the dihedral angles in 3D molecules. It is also promising to use MXMNet as a general tool to learn the representations of molecules in more tasks. Moreover, since molecules can have multiple conformations. It remains unclear how these conformations affect our model and other related GNNs.

\section*{Acknowledgements}
This project has been funded by National Institute of General Medical Sciences (R01GM122845) and National Institute on Aging (R01AD057555) of the National Institute of Health.

\newpage
\putbib[main]
\end{bibunit}

\newpage

\section{Appendix}
\subsection{Dataset Sources}
\paragraph{QM9.}
For the QM9 dataset, we use the source\footnote{\url{https://figshare.com/collections/Quantum_chemistry_structures_and_properties_of_134_kilo_molecules/978904}} provided by~\cite{ramakrishnan2014quantum}. Following the previous works~\cite{gilmer2017neural,schutt2018schnetpack,unke2019physnet,chen2019graph,lu2019molecular}, we process the QM9 dataset by removing about 3k molecules that fail a geometric consistency check or are difficult to converge~\cite{faber2017prediction}. For the properties $U_0$, $U$, $H$, and $G$, only the atomization energies are used by subtracting the atomic reference energies as in~\cite{klicpera_dimenet_2020}. For the property $\Delta \epsilon$, we follow the same way as the DFT calculation and predict it by calculating $\epsilon_{\mathrm{LUMO}}-\epsilon_{\mathrm{HOMO}}$.

\paragraph{PDBBind.} 
For the PDBBind dataset, we use the version\footnote{\url{http://deepchem.io.s3-website-us-west-1.amazonaws.com/datasets/pdbbind_v2015.tar.gz}} that is included in the MoleculeNet benchmark for molecular machine learning~\cite{wu2018moleculenet}. We use log$K_i$ as the target property being predicted, which is proportional to the binding free energy.

\subsection{Baseline Sources}
For the baselines used in the experiment on PDBBind, we find their codes provided by the original papers are based on different frameworks: SchNet~\cite{schutt2018schnetpack} is based on PyTorch~\cite{paszke2019pytorch}, while PhysNet~\cite{unke2019physnet} and DimeNet~\cite{klicpera_dimenet_2020} are based on Tensorflow~\cite{abadi2016tensorflow}. To make fair comparisons and exclude the differences brought by different frameworks, we adopt the implementations of SchNet\footnote{\url{https://github.com/rusty1s/pytorch_geometric/blob/73cfaf7e09/examples/qm9_schnet.py}} and DimeNet\footnote{\url{https://github.com/rusty1s/pytorch_geometric/blob/73cfaf7e09/examples/qm9_dimenet.py}} provided by the widely used PyTorch Geometric library~\cite{fey2019fast} for graph representation learning. Since DimeNet is built based on PhysNet, by comparing their original implementations, we create the implementation of PhysNet based on$^\text{4}$ by changing the corresponding modules. Besides, the code of our MXMNet is also built based on$^\text{4}$.

\subsection{Implementation Details}
For the multi-layer perceptrons (MLPs) used in our MXMNet, they all have 2 layers to take advantage of the approximation capability of MLP~\cite{hornik1989multilayer}. For all activation functions, we use the self-gated Swish activation function~\cite{ramachandran2017searching}. For the basis functions $\boldsymbol{e}_{RBF}$ and $\boldsymbol{a}_{SBF}$, we use $N_{\text{SHBF}}=7$, $N_{\text{SRBF}}=6$ and $N_{\text{RBF}}=16$. To initialize all learnable parameters, we use the default settings used in PyTorch without assigning specific initializations except the initialization for the input node embeddings $\boldsymbol{h}_{g}^{(0)}$ in the global layer: $\boldsymbol{h}_{g}^{(0)}$ are initialized with random values uniformly distributed between $-\sqrt{3}$ and $\sqrt{3}$.

In our experiment on QM9, we use the single-target training following~\cite{klicpera_dimenet_2020} by using a separate model for each target instead of training a single shared model for all targets. The models are optimized by minimizing the mean absolute error (MAE) loss using the Adam optimizer~\cite{kingma2014adam}. We use a linear learning rate warm-up over 1 epoch and an exponential decay with ratio 0.1 every 600 epochs. The model parameter values for validation and test are kept using an exponential moving average with a decay rate of 0.999. To prevent overfitting, we use early stopping on the validation loss. Besides, we repeat our runs 3 times for each MXMNet variant following~\cite{anderson2019cormorant}.

In our experiment on PDBBind, for each model being investigated, we create three weight-sharing, replica networks, one each for predicting the target $G$ of complex, protein pocket, and ligand following~\cite{gomes2017atomic}. The final target is computed by $\Delta G_{\text{complex}} = G_{\text{complex}} - G_{\text{pocket}} - G_{\text{ligand}}$. The full model is trained by minimizing the mean squared error (MSE) loss between $\Delta G_{\text{complex}}$ and the true values using the Adam optimizer~\cite{kingma2014adam}. The learning rate is dropped by a factor of 0.2 every 50 epoch. Moreover, we perform 10-fold cross-validation and repeat the experiments 5 times for each model. The validation losses are used for early stopping.

In Table~\ref{table:hyperparameter}, we list the most important hyperparameters used in our experiments.

\begin{table*}[t]
\caption{List of hyperparameters used in our experiments on QM9 and PDBBind.}
\label{table:hyperparameter}
\small
\centering
\begin{tabular}{lcc}
	\toprule
	\multirow{2}{*}{Hyperparameters} & \multicolumn{2}{c}{Value / Range}\\
	  & QM9 & PDBBind\\
	\midrule
	Batch Size & 32, 128 & 32\\
	Hidden Dim. & 128 & 128\\
	Initial Learning Rate & 1e-3, 1e-4 & 1e-3, 5e-4\\
	Number of Layers & 6 & 2, 3\\
	Max. Number of Epochs & 900 & 250\\
	Dropout & 0 & 0\\
	\bottomrule
\end{tabular}
\end{table*}

\bibliographystyle{unsrt}
\bibliography{appendix}

\end{document}